\documentclass[letterpaper]{article} 
\usepackage{aaai24}  
\usepackage{times}  
\usepackage{helvet}  
\usepackage{courier}  
\usepackage[hyphens]{url}  
\usepackage{graphicx} 
\usepackage{natbib}  
\usepackage{caption} 
\usepackage{algorithm}
\usepackage{algorithmic}
\usepackage{graphicx}
\usepackage{amsmath}
\usepackage{amssymb}
\usepackage{bm}
\usepackage{multirow}
\usepackage{caption}

\usepackage{newfloat}
\usepackage{listings}
\DeclareCaptionStyle{ruled}{labelfont=normalfont,labelsep=colon,strut=off} 
\floatstyle{ruled}
\newfloat{listing}{tb}{lst}{}
\floatname{listing}{Listing}
\title{Enhanced Fine-grained Motion Diffusion for Text-driven Human Motion Synthesis}
\author{
    Dong Wei\textsuperscript{\rm 1}, Xiaoning Sun\textsuperscript{\rm 1}, Huaijiang Sun\textsuperscript{\rm 1}\thanks{Corresponding author.}, Shengxiang Hu\textsuperscript{\rm 1}, Bin Li\textsuperscript{\rm 2}, Weiqing Li\textsuperscript{\rm 1}, Jianfeng Lu\textsuperscript{\rm 1}   \\
}
\affiliations{
    \textsuperscript{\rm 1}School of Computer Science and Engineering, Nanjing University of Science and Technology, Nanjing, China\\
    \textsuperscript{\rm 2}Tianjin AiForward Science and Technology Co., Ltd., Tianjin, China\\
    \{csdwei, sunxiaoning, sunhuaijiang, hushengxiang, li\_weiqing, lujf\}@njust.edu.cn, libin@aiforward.com

}

\usepackage{bibentry}

\begin{document}

\maketitle

\begin{abstract}
The emergence of text-driven motion synthesis technique provides animators with great potential to create efficiently. However, in most cases, textual expressions only contain general and qualitative motion descriptions, while lack fine depiction and sufficient intensity, leading to the synthesized motions that either (a) semantically compliant but uncontrollable over specific pose details, or (b) even deviates from the provided descriptions, bringing animators with undesired cases. In this paper, we propose \emph{DiffKFC}, a conditional diffusion model for text-driven motion synthesis with KeyFrames Collaborated, enabling realistic generation with collaborative and efficient dual-level control: coarse guidance at semantic level, with only few keyframes for direct and fine-grained depiction down to body posture level. Unlike existing inference-editing diffusion models that incorporate conditions without training, our conditional diffusion model is explicitly trained and can fully exploit correlations among texts, keyframes and the diffused target frames. To preserve the control capability of discrete and sparse keyframes, we customize dilated mask attention modules where only partial valid tokens participate in local-to-global attention, indicated by the dilated keyframe mask. Additionally, we develop a simple yet effective smoothness prior, which steers the generated frames towards seamless keyframe transitions at inference. Extensive experiments show that our model not only achieves state-of-the-art performance in terms of semantic fidelity, but more importantly, is able to satisfy animator requirements through fine-grained guidance without tedious labor.
\end{abstract}

\section{Introduction}

Applying human motion synthesis to film or game industries \cite{van2010real} could greatly reduce the dependence on costly motion capture system. How to \emph{efficiently} generate \emph{high-quality} motions \emph{towards animator needs} becomes a key issue. In terms of efficiency, the technique of text-driven motion synthesis has achieved initial success, where textual descriptions are used to provide an overall, coarse guidance at semantic level, under frameworks of GANs \cite{text2act-icra} or VAEs \cite{guochuan-text2mot,temos}, but the synthesized motions often suffer from unrealistic behaviors, such as foot sliding.
\begin{figure}[t]
\centering
\includegraphics[width=0.99\columnwidth]{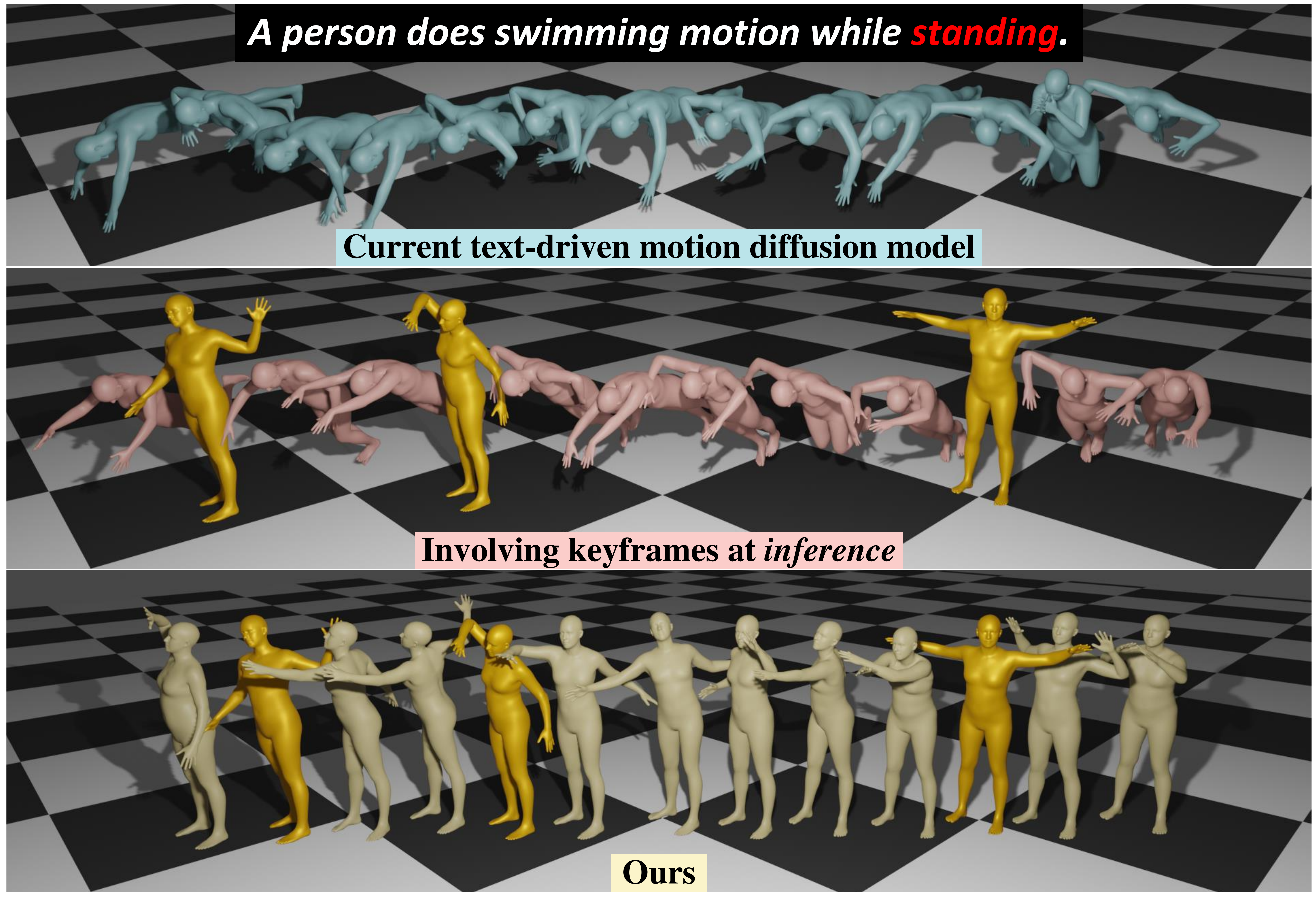}
\caption{Top: Current text-driven motion diffusion models, such as MDM \cite{tevet}, may miss the short \emph{standing} word and generate a total \emph{swimming} motion. Middle: Directly imposing keyframe (golden) conditions at \emph{inference} fails to solve this problem, and results in heavy discontinuities. Bottom: Our collaborative dual-level control paradigm produces the realistic motion towards animator expectations.}
\label{fig1}
\end{figure}

Recent surge of denoising diffusion probabilistic models (DDPMs) has significantly improved the generative quality of text-to-motion synthesis \cite{motiondiffuse,tevet,mofusion}. However, the specific posture details are still uncontrollable for animators. Reaching animator expectations requires more than just semantic fidelity; it involves realizing various intricate details that are difficult to be explicitly expressed using free-form texts, which, unfortunately, are totally ignored in current models. Our research shows that this problem can be effectively solved by incorporating only a few keyframes as direct and fine-grained control, along with coarse textual guidance to establish a collaborative dual-level control paradigm. Additionally, this control paradigm allows for two incidental abilities that are also lacking in current text-to-motion models: correction of semantic misunderstanding and toleration on spelling errors.

Enabling fine-grained control over text-driven motion diffusion models is not trivial. Seemingly, an alternative way is to perform local editing at \emph{inference stage}, which can be roughly divided into two strategies: inpainting \cite{repaint, blended}, which uses keyframes to in-fill the rest frames under the given texts; gradient-based guidance sampling \cite{beat, glide}, which perturbs the reverse-process mean with the gradient of the keyframe guidance. Nevertheless, such naive approaches cannot yield satisfactory results due to the following reasons: In practical scenarios, such as character animation and gaming, the keyframes provided by animators are often distributed in a \emph{scattered} or \emph{discrete} manner, and therefore tend to be perceived as noise by current motion diffusion models during inference. 
In this case, directly imposing keyframes at inference could not bring effective or beneficial impact, but result in huge gap between keyframes and the generated content, to which human eyes are very sensitive. Moreover, this defect may become even more pronounced when semantic misunderstanding happens, as shown in Figure \ref{fig1}.

In this paper, we propose \emph{DiffKFC}, a conditional diffusion model for text-driven motion synthesis with KeyFrames Collaborated. To overcome the limitations caused by naive usage of keyframes, we proactively learn the dual-level collaboration between keyframes and textual guidance from early \emph{training stage}. A carefully re-designed transformer structure is employed to fully exploits the internal correlations among multi-modal tokens of texts, keyframes and the diffused target frames. Considering the reduction of animator labor to ensure production efficiency, our DiffKFC should guarantee the effectiveness of keyframe control even when the provided number is small. Such sparsity of keyframes poses \emph{another} challenge to their usage, as the useful keyframe information is prone to be overwhelmed when interacting with other massive tokens. To this end, we customize efficient Dilated Mask Attention (DMA) modules, which dilate the sparse valid tokens in keyframe mask step by step to ``complete'' the entire mask to be valid, where the scope of attention operation expands from local to global range, so that the keyframe information can gradually expand into the entire temporal dimension.

During inference, DiffKFC utilizes a simple yet effective temporal smoothness prior as backward propagation to encourage seamless transitions near keyframes, which is different from inference-editing diffusion models that solely rely on keyframes rectification. For user-friendly concerns, we extend the idea of classifier-free guidance \cite{ho-classifier}, and reserve a variable for users to adjust the relative importance of fine-grained keyframe control under the given semantic space. Remarkably, our model is also compatible with flexible body part editing, such as re-generating the upper body part while remaining consistency and rationality with the fixed, conditional lower body part.

In summary, our contributions are as follows:
(a) We are the first to utilize the controllable capability of keyframes for text-driven motion synthesis. It exploits full interaction among multi-modal tokens, enabling a collaborative dual-level control paradigm, i.e., coarse semantic guidance, with keyframes for direct yet fine-grained depiction.
(b) We develop dilated mask attention modules to gradually borrow visible useful information from sparse keyframes with local-to-global attention, benefitting its fusion with other tokens. 
(c) We show that, with very few keyframes (only 2\%), our model achieves 41.6\% improvement against state-of-the-art text-to-motion diffusion models in terms of Frechet Inception Distance (FID) on HumanML3D dataset. Meanwhile, it can help produce the exactly expected motions of animators. We also allow two incidental abilities: correcting semantic misunderstanding and tolerating spelling errors.

\section{Related Work}
\subsection{Human Motion Synthesis}
Human motion synthesis aims to generate realistic motion sequences based on given conditions. Various guidance that can be easily offered by users, such as action labels \cite{actor,guochuan-act2mot}, texts \cite{language2pose,ghosh,guochuan-text2mot,temos,shafir2023human,jiang2023motiongpt}, music \cite{ai-dance,rhythm-dance}, historical poses \cite{mao2019learning,sun2023defeenet} have been employed in motion generation, under frameworks of GANs or VAEs. Particularly, \textbf{text-driven motion synthesis} helps animators generate motions with high efficiency, as motion descriptions contained in free-form texts could guide motions towards the target semantics without tedious labor, while suffering from unrealistic behaviors, such as foot sliding and shaky movements. The recent \textbf{motion diffusion models} \cite{motiondiffuse,flame,tevet,mofusion} have alleviated this problem, where texts are encoded by pre-trained CLIP \cite{clip}, to be fused with motion embeddings to yield higher generative quality. Among them, MDM \cite{tevet} is the most \emph{representative} and \emph{influential} model which employs a lightweight transformer-encoder backbone. The newly proposed MLD \cite{executing} further designs a latent diffusion model. GMD \cite{gmd} additionally involves environment constraints into generations, such as obstacle avoidance. PhysDiff \cite{yuanye} considers the natural laws of physics in the diffusion process to alleviate artifacts such as floating and foot sliding.

However, all the above models are only aimed at generating semantically compliant motions, while ignoring the animator requirements for specific posture details or visual effects that beyond textual descriptions. We propose to incorporate sparse yet useful keyframes for direct and fine-grained depiction, along with coarse textual guidance to form a collaborative dual-level control paradigm.

\subsection{Denoising Diffusion Probabilistic Models}
Denoising Diffusion Probabilistic Models (DDPMs) \cite{diffusion,diffusion-ho,score-song} is a kind of generative models inspired by particle diffusion process in thermodynamics, and have gained huge popularity recently for their high generative quality. The basic elements of each sample (e.g., pixels of the image, joints of the human body) can be regarded as heated particles that gradually diffuse towards a complete noisy state, and the models will learn its reverse process (i.e., denoising) to recover the data distribution. Successes have been achieved in many fields, such as image generation \cite{diffusion-ho,beat,repaint}, point cloud generation \cite{diffusion-cloud,diffusion-cloud-iclr}, audio synthesis \cite{ai-dance,rhythm-dance} and human motion prediction \cite{wei2023human} and synthesis \cite{motiondiffuse,flame,tevet,mofusion}.

Concerning the motion synthesis task, \textbf{inference-editing with frame-wise conditions} is allowed in diffusion models, with no need for additional training. MDM \cite{tevet} and FLAME \cite{flame} can support motion in-betweening, by taking the beginning frames and ending frames as additional conditions, and denoising the rest frames with textual guidance. 
However, such approaches cannot be analogically applied to keyframe conditions. Unlike them, which condition on \emph{consecutive} frames at the start and end of a sequence, our DiffKFC considers that the keyframes provided by animators are often distributed in a \emph{scattered} or \emph{discrete} manner. Consequently, these keyframes tend to be perceived as noise, and would gradually lose their control capability during inference. Therefore, we propose to proactively learn dual-level collaboration from training stage to avoid the above defect.

\section{Proposed Method}
\label{Method}

We present an adapted formulation for our task. Existing plain text-driven motion synthesis is aimed at using a free-form textual description, usually a sentence $ \mathcal{C} $, to generate towards the target $ N $-frame motion sequence $ \mathcal{X}=\{\textbf{x}_{1},\textbf{x}_{2},\cdots,\textbf{x}_{N} \} $ with each human pose $ \textbf{x}_i \in \mathbb{R}^{D}$ represented by either joint rotations or positions. We, additionally, incorporate sparse keyframes $ \mathcal{X}^{kf}=\mathcal{X}\odot \textbf{M} $ at training stage, where $ \textbf{M} $ is a binary mask matrix that preserves keyframe values and zeros the target frames (denoted as $ \mathcal{X}^{ta} $) to be generated. Note that mask matrix $\textbf{M}$ for each sequence could be different. 

\begin{figure}
\centering
\includegraphics[width=0.47\textwidth]{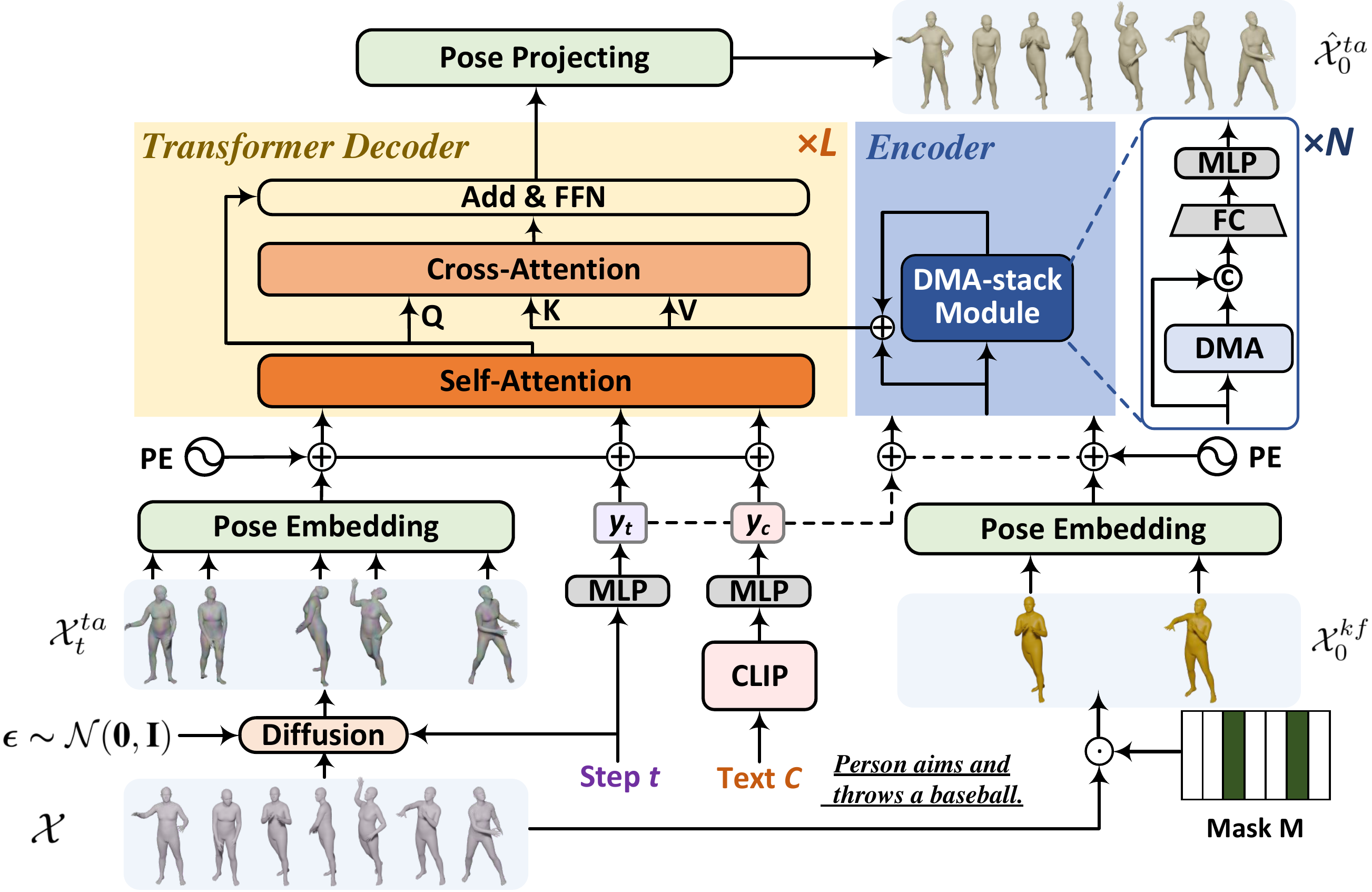} 
\caption{Overview of the proposed DiffKFC method. The encoder of DiffKFC is fed with clean keyframes $\mathcal{X}_{0}^{kf}$, while the decoder of DiffKFC is fed with the target frames $\mathcal{X}_{t}^{ta}$ in a noising step $t$ and the output of the encoder. The diffusion step $t$ and text description $\mathcal{C}$ are projected to the tokens $\textbf{y}_{t}$ and $\textbf{y}_{c}$, which are then combined with input tokens. Dilated Mask Attention (DMA) modules are proposed to extract refined tokens from sparse keyframes and texts to conduct better interaction with encoded diffused motion tokens.}
\label{fig:2}
\end{figure}

\textbf{Overview of Method.} Our objective is to generate motions conformed to the given text, with sparse keyframes controlled towards the expectations of animators. As shown in Figure \ref{fig:2}, we propose a conditional diffusion model with our re-designed transformer structure to generate realistic and controllable human motions. We train our model using classifier-free guidance to enable flexible and convenient adjustment on the relative importance of fine-grained control for animators. To generate smooth transitions near keyframes, we develop a simple yet effective smoothness prior at each denoising step. Our design further brings two incidental benefits of correcting semantic misunderstanding and tolerating spelling errors. 

\subsection{Diffusion Model with Keyframe Collaborated}
Denoising diffusion probabilistic model can be described as a latent variable model with the \textit{forward} diffusion process and the \textit{reverse} diffusion process. Intuitively, the diffusion process gradually injects small amount of noise to the data until it is totally transformed to isotropic Gaussian noise, while the reverse process learns to gradually eliminate the noise to recover the original meaningful data. Recently, diffusion models have been introduced into human motion synthesis due to their powerful generative capabilities. 

Unlike existing motion diffusion models \cite{motiondiffuse,tevet,flame} that are generally designed for text-to-motion generation, while bluntly involving the beginning and ending frames only during \emph{inference} for motion in-betweening, our diffusion model, however, regards keyframes as an important element for fine-grained control \emph{since training stage}, taking them as a conditional input to fully learn their internal information and proactively exploit their delicate correlations with texts.

\textbf{Diffusion Process.} Let $ \mathcal{X}_{0}^{ta} $ denotes the ground truth sequence (with $ N $ poses) that required to be generated. We define the forward diffusion process as a fixed posterior distribution $ q(\mathcal{X}^{ta}_{1:T}|\mathcal{X}^{ta}_{0}) $, which can be modeled as a Markov chain according to a fixed variance schedule $ \beta_{1},\cdots,\beta_{T} $:
\begin{equation}
\label{eq:1}
q(\mathcal{X}^{ta}_{t}|\mathcal{X}^{ta}_{t-1})=\mathcal{N}(\mathcal{X}^{ta}_{t};\sqrt{1-\beta_{t}}\mathcal{X}^{ta}_{t-1},\beta_{t}\textbf{I}).
\end{equation}
A special property is that the diffusion process can be represented in a closed form for any diffusion step $t$.

\textbf{Reverse Process.} Given a set of specified keyframe $ \mathcal{X}^{kf} $ and a text prompt $ \mathcal{C} $, we consider the reverse dynamics of the above diffusion process to recover the meaningful sequence $ \mathcal{X}_{0}^{ta} $ from the white noise $ \mathcal{X}_{T}^{ta} $. We formulate this reverse dynamics with the following parameterization:
\begin{small}
\begin{equation}
\label{eq:2}
p_{\theta}(\mathcal{X}^{ta}_{t-1}|\mathcal{X}^{ta}_{t},\mathcal{X}^{kf}_{0},\mathcal{C})=\mathcal{N}(\mathcal{X}^{ta}_{t-1};\mu_{\theta}(\mathcal{X}^{ta}_{t},\mathcal{X}^{kf}_{0},\mathcal{C},t),\sigma_{t}^{2}\textbf{I}),
\end{equation} 
\end{small}
where $ \mu_{\theta} $ is a neural network to estimate the means, $ \sigma_{t}^{2}\textbf{I} $ is a user-defined variance term of Gaussian transition, and $\mathcal{X}_{0}^{kf}$ denotes the clean (not diffused) keyframe matrix.

Notably, inference-editing diffusion models usually approximate Eq.(\ref{eq:2}) with $ p_{\theta}(\mathcal{X}^{ta}_{t-1}|\mathcal{X}^{ta}_{t},\mathcal{X}^{kf}_{t},\mathcal{C}) $, which would harm useful information in the observed keyframes and mislead the text prompt, as depicted in \cite{csdi}.

\textbf{Loss Function.} 
As revealed in \cite{benny2022dynamic}, the prediction of $\mu_{\theta}$ can be replaced by either the prediction of $\epsilon$, denoted as $\epsilon_\theta$, or the prediction of $\mathcal{X}_{0}$, denoted as $\mathcal{X}_{\theta}$. Here, we select $\mathcal{X}_{\theta}$ model due to the following reasons: First, $\mathcal{X}_{\theta}$ model could have more influence than $\epsilon_\theta$ on the sampling mean when $t$ approximates to 0. This is beneficial for fine-tuning using our smoothness prior at the last denoising steps. Second, for the task of motion generation, geometric losses are necessary to constrain the human poses, encouraging coherent and plausible motions. Therefore, the loss function can be derived as follows:
\begin{small}
\begin{equation}
\label{eq:3}
\mathcal{L}_{simple}=\mathbb{E}_{\mathcal{X}^{ta}_{0},t}\left[\|(\mathcal{X}^{ta}_{0}-\mathcal{X}_{\theta}(\mathcal{X}^{ta}_{t},\mathcal{X}^{kf}_{0},\mathcal{C},t))\odot (\textbf{1}-\textbf{M})\|_{2}^{2}\right],
\end{equation} 
\end{small}
where $\textbf{1}-\textbf{M}$ is a mask that corresponds to $\mathcal{X}^{ta}_{0}$. Similar to \cite{tevet}, we adopt auxiliary kinematic losses $ \mathcal{L}_{phy} $ including joint positions, foot contact and velocities, to enforce physical properties and prevent artifacts like foot sliding. 

\textbf{Transition Guidance.}
The ability to generate natural transitions near keyframes is critical for realism of human motions. However, we find that the motions generated by our keyframes-conditioned model tend to have slight discontinuities near keyframes, which can be easily perceived by human eyes. To mitigate this problem, we leverage the idea of Discrete Cosine Transform (DCT) to define our temporal smoothness prior during inference, motivated by \cite{mao}. More concretely, we concatenate $ l $ generated frames before and after the $ i $-th keyframe $ \textbf{x}^{kf}_{i} $ to form a new sequence $ \textbf{G}_{i}=[\hat{\textbf{x}}_{i-l},\cdots,\hat{\textbf{x}}_{i-1},\textbf{x}^{kf}_{i},\hat{\textbf{x}}_{i+1},\cdots,\hat{\textbf{x}}_{i+l}] $. We remove the high-frequency DCT basis, and approximate this sequence by $ \hat{\textbf{G}}_{i}=\textbf{G}_{i}\textbf{D}\textbf{D}^{T} $, where $\textbf{D}\in \mathbb{R}^{(2l+1)\times m}$ encodes the first $ m $ DCT bases. Our transition loss is defined as:
\begin{equation}
\label{eq:4}
\mathcal{L}_{tr}=\frac{1}{(2l+1)\cdot K}\sum_{i=1}^{K}\|\hat{\textbf{G}}_{i}-\textbf{G}_{i}\|_{2}^{2},
\end{equation} 
where $ K $ denotes the number of keyframes.

To apply this idea to diffusion models, we can replace the classifier with our smoothness prior in classifier guidance \cite{beat}. With guidance scale $r$, we perturb the reverse-process mean using the gradient of the transition loss with respect to keyframes:
\begin{equation}
\hat{\mu}_{\theta}(\mathcal{X}_{t}^{ta}|\mathcal{X}_{0}^{kf})=\mu_{\theta}(\mathcal{X}_{t}^{ta}|\mathcal{X}_{0}^{kf})+r \cdot \Sigma_{\theta}(\mathcal{X}_{t}^{ta}|\mathcal{X}_{0}^{kf})\nabla_{\mathcal{X}_{t}^{ta}}\mathcal{L}_{tr},
\end{equation}

\textbf{Classifier-free Guidance.} 
From the perspective of animators, we wish to adjust the relative importance of fine-grained keyframes $\mathcal{X}_{0}^{kf}$ on the motions against coarse-grained text description $\mathcal{C}$. To this end, we extend the core idea of the classifier-free guidance \cite{ho-classifier}, which suggests interpolating or extrapolating between the conditioned model and the unconditioned model. Then, our classifier-free guidance inference can be implemented as: $\mathcal{X}_{\theta}(\mathcal{X}_{t}^{ta},t,\mathcal{C},\mathcal{X}_{0}^{kf})=\mathcal{X}_{\theta}(\mathcal{X}_{t}^{ta},t,\mathcal{C},\emptyset)+s\cdot (\mathcal{X}_{\theta}(\mathcal{X}_{t}^{ta},t,\mathcal{C},\mathcal{X}_{0}^{kf})-\mathcal{X}_{\theta}(\mathcal{X}_{t}^{ta},t,\mathcal{C}, \emptyset))$, where $s$ controls the intensity of keyframe control in text-driven motion synthesis. For training, our model learns both the dual-level controlled and coarse text-only controlled distributions by randomly dropping keyframes in 10\% of the samples, such that $\mathcal{X}_{\theta}(\mathcal{X}^{ta}_{t},\mathcal{C},t)$ approximates $p(\mathcal{X}_{0}^{ta})$.

\subsection{Network Architecture}
To enable variable-length motion generation and capture global temporal information, we implement the denoising network $\mathcal{X}_{\theta}$ with a re-designed Transformer \cite{transformer} structure, as depicted in Figure \ref{fig:2}. Our model consists of feature extraction, the transformer decoder (orange) and the keyframe encoder (blue). In particular, feature extraction uses MLP to encode the target human poses and the diffused time step. The pre-trained CLIP \cite{clip} model is employed as the text encoder, similar to MDM. Our transformer decoder contains self-attention layers (SA), cross-attention layers (CA) and feed-forward network (FFN). It first models strong global correlation between any two tokens (including diffused target poses $\mathcal{X}_{t}^{ta}$, text $\mathcal{C}$ and diffused step $t$) via SA mechanism. Then, it accepts the outputs of our keyframe encoder to exchange the keyframe information by CA layers, and is then fed into FFN. After the fusion with keyframe tokens, diffused pose tokens pass the learned information to its own branch, which naturally enriches the representation of each diffused pose token.

However, for our keyframe encoder, we find that directly using ViT encoder fails to borrow visible useful information for training. \emph{This is common as given keyframes are sparse that tend to be overwhelmed when fused into CA layers.} 
One possible solution is to use motion in-betweening models (like SLERP \cite{shoemake1985animating}) as a pre-processing step, which interpolates keyframes to form a full sequence, to be fed into CA layers. Nevertheless, this lowers the flexibility of keyframe selection, as it requires that at least the first frame and last frame must be given, and may introduce misleading guidance that harms generative realism and blurs motion details.
That is where our motivation of DMA lies in.

\textbf{Dilated Mask Attention Module.}
Inspired by \cite{mat} for large hole image inpainting, we propose Dilated Mask Attention stack module to handle large number of missing (target) frame tokens (about 95$\%$ tokens are invalid). Our key insight is that, since the keyframe matrix contains important yet temporarily unusable information due to sparsity, if we take these visible sparse keyframes as starting points, and find a certain strategy to ``complete'' the entire sequence, then the ``completed sequence'' would interact with the diffused target frames and texts more effectively.

Suppose that the input of our keyframe encoder is denoted as $\textbf{Z}\in \mathbb{R}^{d\times N}$. The attention of DMA can be derived by:
\begin{equation}
\label{eq:7}
\text{Att}(\textbf{Q},\textbf{K},\textbf{V})=\text{Softmax}(\frac{\textbf{Q}\textbf{K}^{T}+\textbf{M}^{'}}{\sqrt{d}})\textbf{V},
\end{equation}
where $\textbf{Q}$, $\textbf{K}$, $\textbf{V}$ are the learnable query, key, value matrices, respectively. The mask $\textbf{M}^{'}\in \mathbb{R}^{N\times N}$ is defined as:
\begin{equation}
\label{eq:8}
\textbf{M}^{'}_{ij}=\left\{
   \begin{array}{cc}
   0,      & \textrm{if token $j$ is valid,} \\
   -\infty , & \textrm{if token $j$ is invalid,}
   \end{array}
\right.
\end{equation}
where only valid tokens participate in computing weights, while invalid tokens are ignored in current state.

\begin{figure}
\centering
\includegraphics[width=0.47\textwidth]{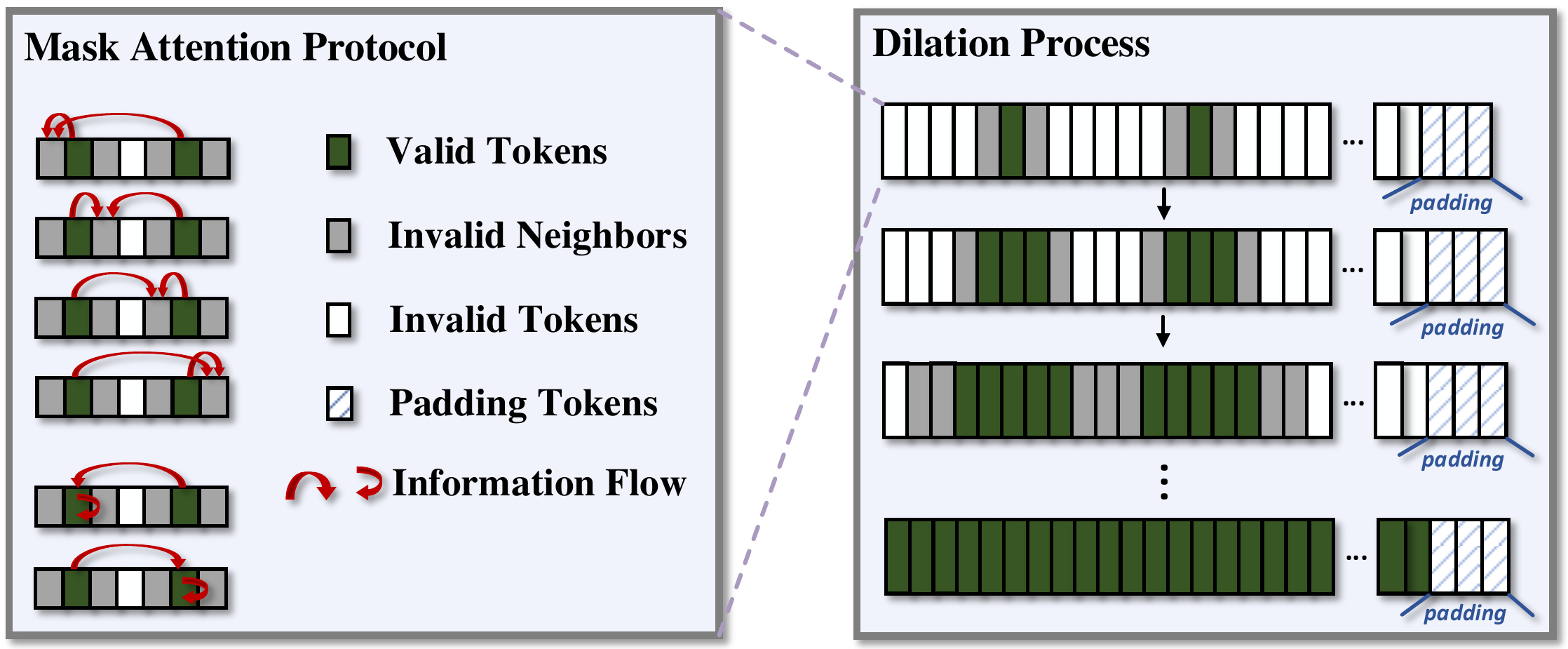} 
\caption{Illustration of dilation strategy. Left: Mask attention protocol. In the range of valid tokens and their invalid neighbors, the output of attention is computed as the weighted sum of valid tokens while invalid tokens are ignored. Right: Dilation process. The invalid neighbors are automatically updated to be valid through each attention. Finally, the whole sequence becomes fully valid except padding tokens (no benefit to interact with diffused tokens).}
\label{fig:3}
\end{figure}

The mask $\textbf{M}^{'}$ indicates whether a token is valid, and it is initialized by the input keyframe mask $\textbf{M}$. We design a valid-token-dilation strategy, which automatically activates invalid neighbors to become valid step by step. As illustrated in Figure \ref{fig:3}, at every dilation step, attention operation is conducted between every \textbf{(valid token, invalid neighbor)} pair, and therefore dilates the valid region. After several times of such operation, the whole sequence is updated to be fully valid (except padding region). Our dilation on valid regions is a gradual process, which forces keyframe information to gradually expand to the entire temporal dimension, with the following advantages: (a) benefitting full interaction in CA layers with tokens of diffused frames and texts; (b) alleviating the problems caused by keyframe sparsity, which helps to generate more natural and detailed motions.

In practice, we remove layer normalization to reduce the importance of invalid tokens, and take advantage of feature concatenation to replace residual learning similar to popular diffusion ViT backbone \cite{bao2022all}. The input and output of attention are concatenated to be fed into a fully-connected layer:
\begin{equation}
\label{eq:9}
\textbf{Z}^{'}_{k}=\mathsf{FC}([\mathsf{DMA}(\textbf{Z}_{k-1}) \Vert \textbf{Z}_{k-1}]),\quad
\textbf{Z}_{k}=\mathsf{MLP}(\textbf{Z}^{'}_{k}),
\end{equation}
where $\textbf{Z}_{k}$ is the output of the $k$-th DMA block. A global skip connection is also adopted to prevent gradient exploding.

\section{Experiments}
In this section, to evaluate our proposed model, we introduce datasets, evaluation metrics, implementation details, and comparable baselines. Results, visualized comparisons with discussion are followed. Ablation study is conducted to show the impact of each component. 

\begin{table*}[ht]
\centering
\scalebox{0.92}{
\begin{tabular}{l|ccc|ccc}
\hline
& \multicolumn{3}{c|}{HumanML3D} & \multicolumn{3}{c}{KIT} \\ \cline{2-7} 
& FID $\downarrow$ & \begin{tabular}[c]{@{}c@{}}R-Precision\\ (Top-3) $\uparrow$\end{tabular} & Diversity $\to$ & FID $\downarrow$ & \begin{tabular}[c]{@{}c@{}}R-Precision\\ (Top-3) $\uparrow$\end{tabular} & Diversity $\to$ \\ \hline
Real motion & $0.002^{\pm.000}$ & $0.797^{\pm.002}$ & $9.503^{\pm.065}$ & $0.031^{\pm.004}$ & $0.779^{\pm.006}$ & $11.08^{\pm.097}$ \\ \hline
JL2P(3DV'19) & $11.02^{\pm.046}$ & $0.486^{\pm.002}$ & $7.676^{\pm.058}$ & $6.545^{\pm.072}$ & $0.483^{\pm.005}$ & $9.073^{\pm.100}$ \\
Hier(ICCV'21) & $6.532^{\pm.024}$ & $0.552^{\pm.004}$ & $8.332^{\pm.042}$ & $5.203^{\pm.107}$ & $0.531^{\pm.007}$ & $9.563^{\pm.072}$ \\
T2M(CVPR'22) & $1.067^{\pm.002}$ & $0.740^{\pm.003}$ & $9.188^{\pm.002}$ & $2.770^{\pm.109}$ & $0.693^{\pm.007}$ & $10.91^{\pm.119}$ \\
MoFusion(CVPR'23) & - & $0.492$ & $8.82$ & - & - & - \\
MLD(CVPR'23) & $0.473^{\pm.013}$ & $\bf{0.772}^{\pm.002}$ & $9.724^{\pm.082}$ & $0.404^{\pm.027}$ & $\bf{0.734}^{\pm.007}$ & $10.80^{\pm.117}$ \\
PhysDiff(ICCV'23) & $0.433$ & $0.631$ & - & - & - & - \\ 
GMD(ICCV'23) & $0.212$ & $0.670$ & $9.440$ & - & - & - \\ 
MDM(ICLR'23) & $0.544^{\pm.044}$ & $0.611^{\pm.007}$ & $9.559^{\pm.086}$ & $0.497^{\pm.021}$ & $0.396^{\pm.004}$ & $10.85^{\pm.109}$ \\\hline
DiffKFC & $\bf{0.111}^{\pm.031}$ & $0.686^{\pm.006}$ & $\bf{9.498}^{\pm.082}$ & $\bf{0.164}^{\pm.026}$ & $0.420^{\pm.007}$ & $\bf{10.98}^{\pm.108}$ \\ \hline
\end{tabular}}
\caption{Results of baselines and DiffKFC (with keyframe rate 5\%) on HumanML3D and KIT datasets. $\to$ means results are better when closer to that of real motion. We evaluate with 20 times of running for each metric, under 95\% confidence interval. \textbf{Bold} indicates best results; ``-'' means unavailable results.}
\label{tab1}
\end{table*}

\subsection{Datasets \& Evaluation Metrics}

\noindent\textbf{KIT Motion-Language dataset} \cite{kit} is a text-to-motion dataset that contains 3,911 human motion sequences and 6,353 sentences of textual descriptions. Dataset split procedure is consistent with prior \cite{language2pose,temos,flame}.

\noindent\textbf{HumanML3D} \cite{guochuan-text2mot} is a new dataset that annotates existing 3D motion capture datasets AMASS \cite{amass} and HumanAct12 \cite{guochuan-act2mot}, containing 14,616 motions with 44,970 textual descriptions. Each frame is represented by the concatenation of root velocities, joint positions, joint velocities, joint rotations, and foot contact labels. We follow \cite{guochuan-text2mot,tevet} to use this representation as well as on KIT dataset.

\noindent\textbf{Evaluation Metrics.}
We use the following matrices of text-to-motion task provided by \cite{guochuan-text2mot}. 
(1) \emph{Frechet Inception Distance} (\emph{FID}): similarity between the distribution of generations and ground truth.
(2) \emph{R-Precision}: the alignment between the generated motion and corresponding text.
(3) \emph{Diversity}: the average body joint differences between each pairs of generated motions that are randomly split.

Particularly, to evaluate our distinctive fine-grained control property, we further employ (4) \emph{Average Displacement Error (ADE)}: the average discrepancies over all timesteps (keyframe excluded) between the desired motion (i.e., GT) and the generated motion closest to GT, which is a metric borrowed from \cite{dlow} for diverse human motion prediction. We design (5) \emph{Keyframe Generative Error (K-Err)}: the average discrepancies between the keyframe and the corresponding generated frame. (6) \emph{Keyframe Transition Smoothness (K-TranS)}: the average smoothness between the keyframe and its adjacent generated frames.

\subsection{Baselines \& Implementation Details}

\noindent\textbf{Baselines.}
We compare the performance of our model with text-to-motion baselines: JL2P \cite{language2pose}, Hier \cite{ghosh}, T2M \cite{guochuan-text2mot}, MDM \cite{tevet}, MoFusion \cite{mofusion}, MLD \cite{executing}, GMD \cite{gmd} and PhysDiff \cite{yuanye}, with the last four models being motion diffusion models.

\noindent\textbf{Implementation Details.}
As for the diffusion model, the number of diffusion steps $t$ is 1,000 with cosine beta scheduling following \cite{tevet,flame}. We employ a frozen \emph{CLIP-ViT-B/32} \cite{clip} model to encode the text prompt $\mathcal{C}$. For each motion sequence, we select 5$\%$ frames as keyframes for fine-grained control, and at least 1 frame will be chosen as keyframe. The encoder of our DiffKFC has 8 layers, and their dilated step size (the number of invalid neighbors of each valid token) is set to $\{2,2,4,4,6,6,8,N\}$. As for the decoder, we build up 8 transformer layers with 8 heads, latent dimension as 512 and feed-forward size 1,024. During inference, the transition guidance scale $r$ and classifier-free guidance scale $s$ are set as 100.0 and 2.5, respectively. We use Adam optimizer with learning rate set to 0.0001. Our model is trained under Pytorch framework using NVIDIA RTX 3090, with batch size 64 for 500K steps on HumanML3D and 200K steps on KIT.

\begin{figure*}
\centering
\includegraphics[width=0.98\textwidth]{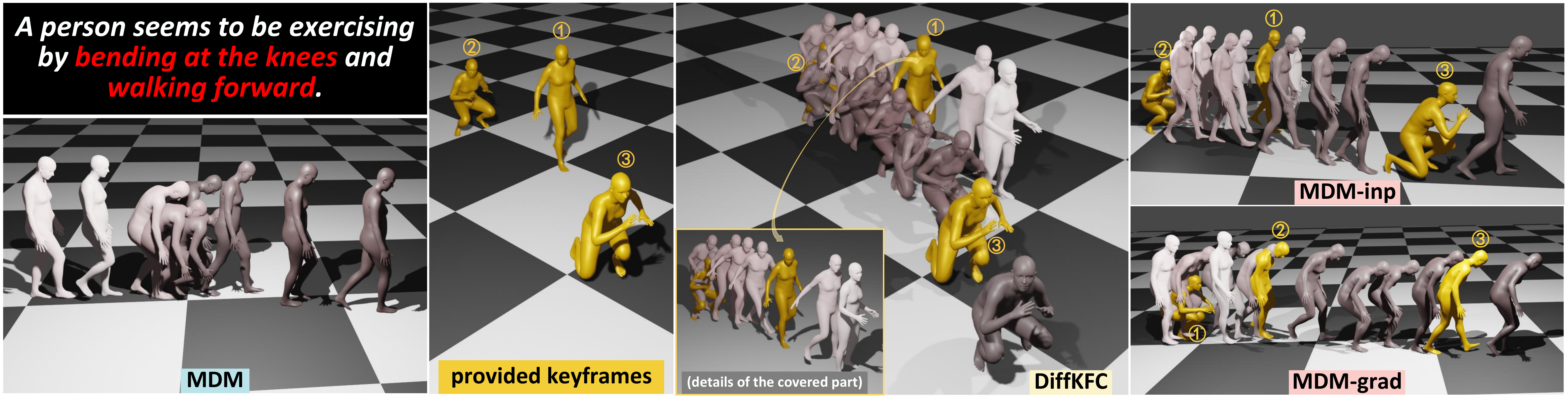} 
\caption{The description is ambiguous on whether ``bending \emph{then} walking'' or ``bending \emph{while} walking''. MDM generates motions that compliant to the former semantics, but obviously do not meet animator needs (reflected in GT keyframes). Our DiffKFC not only removes this ambiguity (identifying the latter semantics), but also generates coherent and keyframe-aligned motions. Visualization of naive inference-editing approaches are also given. MDM-inp yields a continuous walking motion interspersed with single-frame of sudden knee bending, which is a typical unrealistic phenomenon; while for MDM-grad, the generated pose at keyframe positions are different from real keyframes, indicating its failure towards desired visual effects.
}
\label{bending}
\end{figure*}

\subsection{Results}

\noindent\textbf{Comparisons with plain text-driven baselines.}
Table \ref{tab1} shows the results of DiffKFC and baselines, which prove the efficacy of incorporating keyframes. Our generative quality surpasses other models (including the most advanced motion diffusion models) by a significant margin,
with \emph{FID} obtaining 74.4\% improvement over SOTA PhysDiff \cite{yuanye} on HumanML3D, and 59.4\% improvement over SOTA MLD \cite{executing} on KIT, under keyframe rate 5\% of fine-grained control. Our \emph{Diversity} is the closest to that of real motions, which means that our generations are more aligned with the natural law of human behaviors. Our method also achieves the comparable performance with baselines in terms of \emph{R-Precision}. 

\noindent\textbf{Comparisons of fine-grained control property.}
To demonstrate the effectiveness of DiffKFC in enabling our distinctive fine-grained control, firstly, we need to prove the necessity of introducing keyframes into \emph{diffusion training stage}, while existing text-driven motion diffusion baseline cannot acquire this control capability by \emph{naive inference-editing}. Secondly, we need to prove that even if existing baseline has keyframes participating in \emph{training}, simple fusion of texts and keyframes still fails to yield satisfactory improvements. \emph{Only when both aspects above are proven can we conclude that DiffKFC is meaningful in enabling keyframe control to meet animator expectations.}

\begin{table}[t]
\centering
\scalebox{0.73}{
\begin{tabular}{l|ccccc}
\hline
& FID $\downarrow$ & \begin{tabular}[c]{@{}c@{}}R-Precision\\ (Top-3) $\uparrow$\end{tabular} & ADE $\downarrow$ & K-Err $\downarrow$ & K-TranS $\to$ \\ \hline
Real motion & $0.002$ & $0.797$ & - & - & $0.125$ \\ \hline
MDM & $0.544$ & $0.611$ & $1.868$ & $1.826$ & $1.131$ \\
MDM-inp & $0.386$ & $0.642$ & $1.498$ & - & $0.847$ \\
MDM-grad & $0.449$ & $0.625$ & $1.674$ & $1.629$ & $0.914$ \\
MDM-fus & $0.308$ & $0.658$ & $0.343$ & $0.219$ & $0.254$ \\ \hline
DiffKFC w/o TG & $0.119$ & $\bf{0.686}$ & $\bf{0.198}$ & $\bf{0.037}$ & $0.161$ \\
DiffKFC & $\bf{0.111}$ & $\bf{0.686}$ & $0.205$ & $0.041$ & $\bf{0.136}$ \\ \hline
\end{tabular}}
\caption{Results of MDM with different keyframe using and our DiffKFC on HumanML3D. Keyframe rate 5\% is for all.}
\label{tab2}
\end{table}

Since we build up our model on the basis of MDM transformer layers, we choose MDM as the baseline to conduct the following experiments to compare with.
(a) \textbf{MDM-inp}, MDM inpainting at \emph{inference}, which uses keyframes to in-fill the missing frames with textual guidance; (b) \textbf{MDM-grad}, MDM with gradient-based guidance sampling at \emph{inference}, which perturbs the reverse-process mean with the gradient of keyframe guidance, under the given texts; (c) \textbf{MDM-fus}, MDM with text-keyframe simple fusion at \emph{training}, which uses state-of-the-art motion interpolation model \cite{delta} to interpolate keyframes into a full sequence, and operates cross-attention between this sequence and text directly for condition fusion. In table \ref{tab2}, we show these comparisons with ours.

\begin{table}[t]
\centering
\scalebox{0.77}{
\begin{tabular}{l|ccccc}
\hline
& FID $\downarrow$ & \begin{tabular}[c]{@{}c@{}}R-Precision\\ (Top-3) $\uparrow$\end{tabular} & Diversity $\to$ & ADE $\downarrow$ &K-Err $\downarrow$ \\ \hline
Vanilla Enc. & $0.477$ & $0.620$ & $8.957$ & $1.212$ & $0.715$ \\
Unified Enc. & $0.293$ & $0.655$ & $9.150$ & $0.476$ & $0.181$ \\
DiffKFC & $\bf{0.111}$ & $\bf{0.686}$ & $\bf{9.498}$ & $\bf{0.205}$ & $\bf{0.041}$ \\ \hline
\end{tabular}}
\caption{Ablation studies of network architecture designs. `Enc.' is the abbreviation of the keyframe encoder.}
\label{ab1}
\end{table}

\begin{figure}
\centering
\includegraphics[width=0.99\columnwidth]{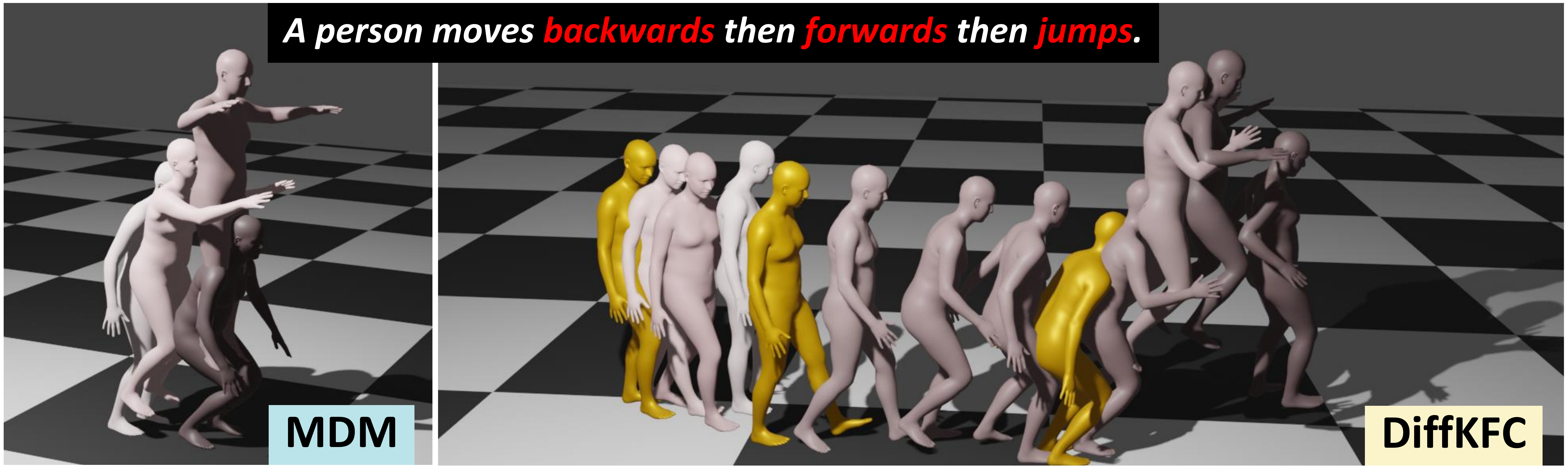} 
\caption{The motion generated by MDM is semantically incorrect (a total \emph{jump} action), but our DiffKFC corrects this misunderstanding and generates the desired motion.}
\label{incidental1}
\end{figure}

Apart from the best \emph{FID} and \emph{R-Precision}, DiffKFC significantly outperforms different keyframe using on MDM in terms of \emph{ADE}, \emph{K-Err} and \emph{K-TranS}. It means our generations are very close to the exactly expected motions, with natural keyframe transitions, which implies that our design is valid for fine-grained control towards animator needs.

Additionally, we remove transition guidance in our DiffKFC (i.e., w/o TG), and get worse \emph{K-TranS}, which proves that this strategy is helpful to generate smoother transitions.

\noindent\textbf{Visualization.}
In Figure \ref{bending}, our DiffKFC produces motions more than just realistic and semantically compliant; it ensures that the generations precisely adhere to the given keyframes, where animator requirements for specific posture details are predominantly reflected. However, this cannot be realized by naive inference-editing approaches. MDM-inp fails to exert effective fine-grained control over large number of missing frames, leading to undesired motion interspersed with unrealistic sudden pose change. MDM-grad can generate coherent motions, but the generated pose at keyframe positions are different from real keyframes, making the whole sequence less likely to meet the expectations.

\begin{figure}
\centering
\includegraphics[width=0.99\columnwidth]{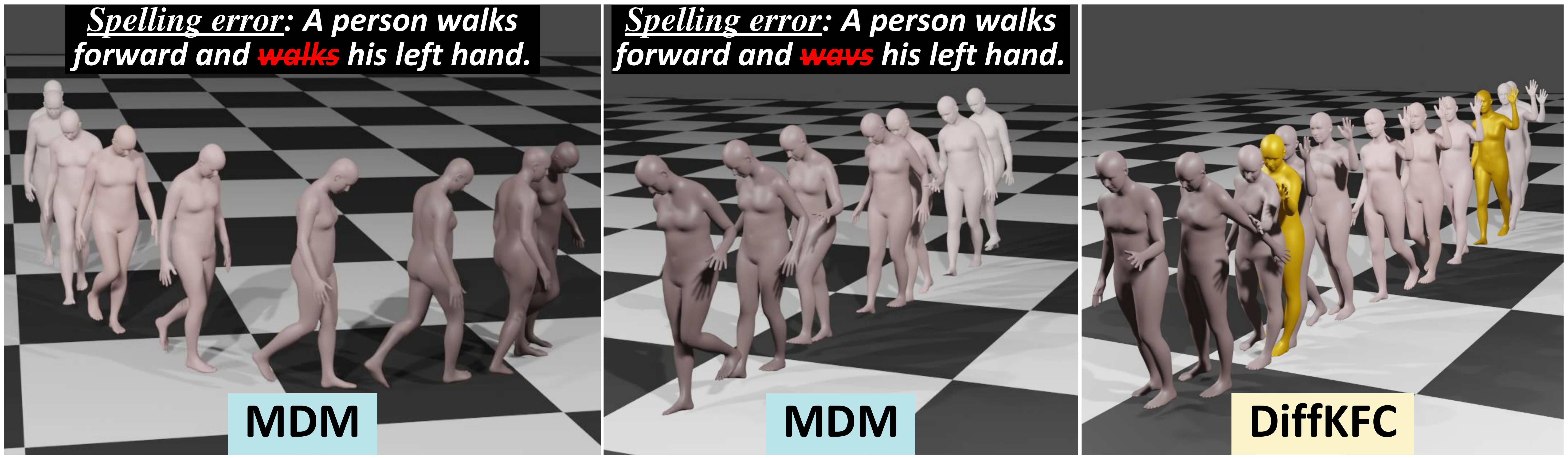} 
\caption{True description: ``\emph{A person walks forward and \textbf{waves} his left hand.}'' When spelling \emph{waves} mistakenly into \emph{walks} or \emph{wavs}, MDM fails to extract the information and causes different failures, but our DiffKFC can tolerate both of these errors and produce the desired motion.}
\label{incidental2}
\end{figure}

We also show visualizations of correcting semantic misunderstanding and tolerating spelling errors in Figure \ref{incidental1} and \ref{incidental2}, where our DiffKFC constantly outperforms the baseline. Notably, in all our figures, darker colors of human body indicate later occurrence.

\subsection{Ablation Study}

\noindent\textbf{Discussion on the effectiveness of DMA.}
We discuss its effectiveness by conducting the following two experiments on HumanML3D, shown in Table \ref{ab1}.
(a) We substitute our DMA-stack structure with a vanilla transformer encoder (i.e., Vanilla Enc.). Results are worse, as the information in sparse keyframes is overwhelmed when fused into CA layers, indicating that our dilation strategy is valid;
(b) For conditional diffusion model, a common design for condition encoding is to directly concatenate the condition part and the diffused target part, which are then together fed into a unified encoder (i.e., Unified Enc.). However, such unified encoding would make it hard to distinguish between keyframes and noised targets, which harms the useful information and leads to unsatisfactory results. In our DiffKFC, we separately encodes keyframes and diffused target frames, with the specially designed DMA for keyframe encoding, and therefore avoid the above problem.

\begin{table}[t]
\centering
\scalebox{0.77}{
\begin{tabular}{l|ccccc}
\hline
\begin{tabular}[l]{@{}l@{}}Keyframe\\ rate \end{tabular} & FID $\downarrow$ & \begin{tabular}[c]{@{}c@{}}R-Precision\\ (Top-3) $\uparrow$\end{tabular} & Diversity $\to$ & ADE $\downarrow$ & K-Err $\downarrow$ \\ \hline
10\% & $\bf{0.079}$  & $\bf{0.694}$  & $\bf{9.500}$ & $\bf{0.124}$ & $\bf{0.039}$ \\
5\% & $0.111$ & $0.686$ & $9.498$ & $0.205$ & $0.041$ \\
2\% & $0.253$ & $0.665$ & $9.441$ & $0.387$ & $0.050$ \\
0\% & $0.597$ & $0.602$ & $9.576$ & $1.831$ & $1.829$ \\ \hline
\end{tabular}}
\caption{Comparison of DiffKFC with different keyframe rates on HumanML3D dataset.}
\label{ab2}
\end{table}

\noindent\textbf{Discuss on keyframe rate.}
Compared to our keyframe number 5\% of the total frames, we further set it to 10\%, 2\% and 0\%, respectively, to evaluate their impacts. In Table \ref{ab2}, results of 0\% keyframe deteriorate significantly (i.e., plain text-driven). However, when keyframes are involved, results of every setting all get improved. Although the more keyframes the better the results, it still proves that only few keyframes are sufficient to help yield satisfactory performance, which allows reduction on the dependence of dense drawing by animators in real-world scenarios.

\section{Conclusion}

In this paper, we propose DiffKFC, a conditional diffusion model for text-driven motion synthesis with keyframes collaborated. We carefully re-design a transformer structure to allow full interaction among multi-modal tokens, enabling realistic yet efficient generation under dual-level control paradigm. We customize dilated mask attention modules that gradually borrow visible useful information with local-to-global attention to overcome the challenge caused by keyframe sparsity. We additionally find a simple smooth prior to generate smooth transitions during inference. Our model not only achieves state-of-the-art performance in semantic fidelity, but also generates motions that satisfy animator expectations on posture details.

\noindent\textbf{Limitation.}
The inference speed of our DiffKFC is slightly lower than SOTA models due to the encodings of keyframes and about 1,000 reverse steps. Future work could speed up the model with fundamental advances in diffusion models.

\section{Acknowledgements}
This work was supported in part by the National Natural Science Foundation of China (NO. 62176125, 61772272).

\bibliography{aaai24}

\end{document}